\documentclass[sigconf, balance=false]{acmart}

\copyrightyear{2024}
\acmYear{2024}
\setcopyright{acmlicensed}\acmConference[SIGGRAPH Conference Papers '24]{Special Interest Group on Computer Graphics and Interactive Techniques Conference Conference Papers '24}{July 27-August 1, 2024}{Denver, CO, USA}
\acmBooktitle{Special Interest Group on Computer Graphics and Interactive Techniques Conference Conference Papers '24 (SIGGRAPH Conference Papers '24), July 27-August 1, 2024, Denver, CO, USA}
\acmDOI{10.1145/3641519.3657395}
\acmISBN{979-8-4007-0525-0/24/07}

\ccsdesc[500]{Computing methodologies~Computer graphics}
\ccsdesc[500]{Computing methodologies~Modeling and simulation}
\ccsdesc[500]{Computing methodologies~Neural networks}

\acmSubmissionID{175}

\usepackage{booktabs} 

\usepackage{listings}
\usepackage[outline]{contour}
\contourlength{0.1em}%
\usepackage{contour}
\usepackage{balance}
\usepackage{enumitem}
\setlist{leftmargin=5.5mm}

\usepackage{hyperref}       
\hypersetup{
    colorlinks=True,
    citecolor=blue,
    linkcolor=blue,
}
\usepackage{url}
\usepackage{booktabs}
\usepackage{wrapfig}
\usepackage{multirow}

\usepackage{dsfont}

\def\bb{\mathbf{b}}
\def\cc{\mathbf{c}}

\def\hh{\mathbf{h}}

\def\uu{\mathbf{u}}

\def\xx{\mathbf{x}}

\def\zz{\mathbf{z}}

\def\WW{\mathbf{W}}

\def\lL{\mathcal{L}}

\def\uU{\mathcal{U}}

\def\Ee{\mathbb{E}}

\def\Re{\mathbb{R}}

\def\Ve{\mathbb{V}}

\DeclareMathOperator*{\argmin}{arg\,min}

\newcommand\paren[1]{\left(#1\right)}

\newcommand\norm[1]{\left\lVert#1\right\rVert}

\DeclareMathSymbol{@}{\mathord}{letters}{"3B}

\def\latex/{\LaTeX}
\def\bibtex/{\hologo{BibTeX}}

\newcommand{\RN}[1]{%
  \textup{\uppercase\expandafter{\romannumeral#1}}%
}

\newcommand{\ie}{\textit{i.e.}}
\newcommand{\eg}{\textit{e.g.}}

\begin{document}

\title{Neural Control Variates with Automatic Integration}

\author{Zilu Li}
\authornote{Equal Contribution.}
\affiliation{%
 \institution{Cornell University}
  \city{Ithaca}
 \country{USA}
}
\email{zl327@cornell.edu}

\author{Guandao Yang}
\authornotemark[1]
\affiliation{%
 \institution{Stanford University}
  \city{Palo Alto}
 \country{USA}
}
\email{guandao@stanford.edu}

\author{Qingqing Zhao}
\affiliation{%
 \institution{Stanford University}
  \city{Palo Alto}
 \country{USA}
}
\email{cyanzhao@stanford.edu}

\author{Xi Deng}
\affiliation{%
 \institution{Cornell University}
 \city{Ithaca}
 \country{USA}
}
\email{xd93@cornell.edu}

\author{Leonidas Guibas}
\affiliation{%
 \institution{Stanford University}
 \city{Palo Alto}
 \country{USA}
}
\email{guibas@cs.stanford.edu}

\author{Bharath Hariharan}
\affiliation{%
 \institution{Cornell University}
 \city{Ithaca}
 \country{USA}
 }
\email{bharathh@cs.cornell.edu}

\author{Gordon Wetzstein}
\affiliation{%
 \institution{Stanford University}
 \city{Palo Alto}
 \country{USA}
 }
\email{gordon.wetzstein@stanford.edu}

\renewcommand{\shortauthors}{Li \textit{et al.} 2024}

\begin{abstract}
This paper presents a method to leverage arbitrary neural network architecture for control variates.
Control variates are crucial in reducing the variance of Monte Carlo integration, but they hinge on finding a function that both correlates with the integrand and has a known analytical integral. 
Traditional approaches rely on heuristics to choose this function, which might not be expressive enough to correlate well with the integrand.
Recent research alleviates this issue by modeling the integrands with a learnable parametric model, such as a neural network.
However, the challenge remains in creating an expressive parametric model with a known analytical integral.
This paper proposes a novel approach to construct learnable parametric control variates functions from arbitrary neural network architectures.
Instead of using a network to approximate the integrand directly, we employ the network to approximate the anti-derivative of the integrand.
This allows us to use automatic differentiation to create a function whose integration can be constructed by the antiderivative network.
We apply our method to solve partial differential equations using the Walk-on-sphere algorithm~\cite{WoS}.
Our results indicate that this approach is unbiased using various network architectures and achieves lower variance than other control variate methods.
\end{abstract}

\keywords{Control Variates, Monte Carlo Methods, PDE Solvers}
\begin{teaserfigure}
    \vspace{-1.5em}
    \includegraphics[width=\linewidth]{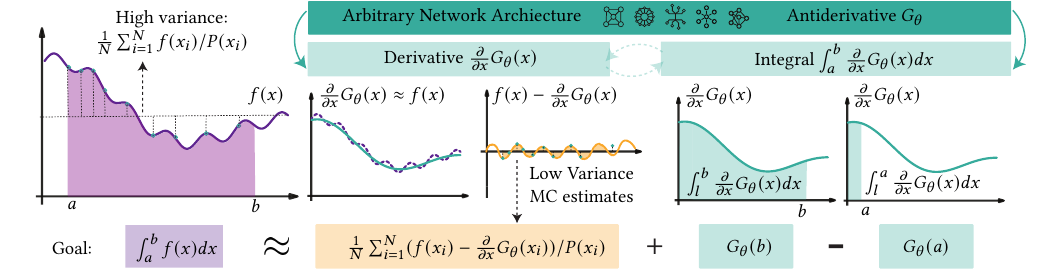}
    \vspace{-2em}
     \caption{We propose a novel method to use arbitrary neural network architectures as control variates (CV). Instead of using the network to approximate the integrand, we deploy it to approximate the antiderivative of the integrand. This allows us to construct pairs of networks where one is the analytical integral of the other, tackling a main challenge of neural CV methods.
    }
    \label{fig:teaser}
\end{teaserfigure}

\maketitle
\section{Introduction}

Monte Carlo (MC) integration uses random samples to estimate the value of an integral. 
It is an essential tool in many computer graphics applications, including solving partial differential equations without discretization ~\cite{WoS} and rendering physically realistic images via ray tracing~\cite{veach1998robust}.
While MC integration provides unbiased estimation for complicated integrals, it suffers from high variance.
As a result, MC integration usually requires a significant amount of samples to produce an accurate estimate.

One common technique to reduce variance is control variates(CV).
The key idea of control variates is to construct an alternative integral that have lower variance.
For example, if we want to integrate a one-dimensional real value function $f:\Re\to\Re$, control variates leverage the following identity to construct different integral:
\begin{align}
    \int_l^u f(x)dx = G + \int_l^u f(x) - g(x) dx,
\end{align}
where $l,u$ defines the integration domain in and $G$ is the integral of real-value function $g : \Re \to \Re$ (\ie~$G = \int_\Omega g(x) dx$).
If the integrand $f-g$ has less variance compared to the integrated $f$, then the right-hand side of this identity can result in an estimator that requires fewer samples to reach the same accuracy.

The key challenge of control variate is finding the appropriate $g$ with known integral while minimizing the variance of $f - g$ under a certain sampling strategy.
Traditional methods try to define the control variates $g$ heuristically, such as by picking a part of $f$ with a known integral. 
These heuristically defined control variates may not correlate with the integrand $f$, limiting their performance.
Recent research has proposed to parameterize the control variate using a learnable function $g_\theta$ and learn the parameter $\theta$ from samples of the integrands $f$~\cite{salaun2022regression,muller2020ncv}.
The hope is to find $\theta$ such that $g_\theta$ can closely match the shape of $f$, making $f - g_\theta$ low variance.
Constructing an expressive parametric function $g_\theta$ with a known integral for all $\theta$, however, remains challenging.
As a result, existing works have limited network architecture choices, such as sum of simple basic functions with known integral~\cite{salaun2022regression} or special neural network architectures such as normalizing flows~\cite{muller2020ncv}.

In this work, we propose a novel method to construct learnable control variate function $g$ from almost arbitrary neural network architectures.
Inspired by neural integration methods~\cite{lindell2021autoint}, instead of using the network to model the control variate $g$ directly, our method defines a network $G_\theta: \Re\to\Re$ to model the anti-derivative of $g$ such that $\frac{\partial}{\partial x}G_\theta(x) = g(x)$. 
By the first fundamental theorem of calculus, we have:
\begin{align}
    G_\theta(u) - G_\theta(l) = \int_l^u \frac{\partial}{\partial x}G_\theta(x) dx.
\end{align}
This allows us to construct a learnable control variate using automatic differentiation frameworks in the following way:
\begin{align}
    \int_l^u f(x) = G_\theta(u) - G_\theta(l) + \int_l^u f(x) - \frac{\partial}{\partial x}G_\theta(x) dx.
\end{align}
Since $\frac{\partial}{\partial x}G_\theta(x)$ is just another neural network, we can use gradient based optimizer to find $\theta$ that minimizes the variance of the integrand $ f(x) -  \frac{\partial}{\partial x}G_\theta(x)$.
This method allows us to use an arbitrary network architecture in place of $G_\theta$, which enables a larger class of parametric functions to be useful for control variates.
We hypothesize that this rich design space contains pairs of $G_\theta$ and $\frac{\partial}{\partial x}G_\theta(x)$ that are expressive and numerically stable enough to match $f$ closely for various problems.

This paper takes the first steps to apply the abovementioned idea to reduce the variance of Monte Carlo integrations in computer graphics applications.
To achieve this, we first extend the neural integration method from \citet{lindell2021autoint} from line integral to spatial integral with different domains, such as 2D disk and 3D sphere.
Directly optimizing these networks to match the integrand can be numerically unstable.
To alleviate this issue, we propose a numerically stable neural control variates estimator and provide corresponding training objectives to allow stable training.
Finally, many graphics applications require solving recursive integration equations where different space locations have different integrand functions.
We modulate the neural networks with spatially varying feature vectors to address this issue.
We apply our method to create control variates for Walk-on-Sphere (WoS) algorithms~\cite{WoS}, which solve PDEs using Monte Carlo integration.
Preliminary results show that our method can provide unbiased estimation from various network architectures.
Our method can produce estimators with the lower variance than all baselines.
To summarize, our paper has the following contributions:
\begin{itemize}
    \item We propose a novel method to use neural networks with arbitrary architecture as a control variate function. 
    \item We propose a numerically stable way to construct control variate estimators for different integration domains.
    \item We demonstrate the effectiveness of our method in solving Laplace and Poisson equations using WoS. 
    Our method can outperform baselines in all settings.
\end{itemize}
\vspace{-1em}
\begin{figure*}
    \centering
    \includegraphics[width=\linewidth]{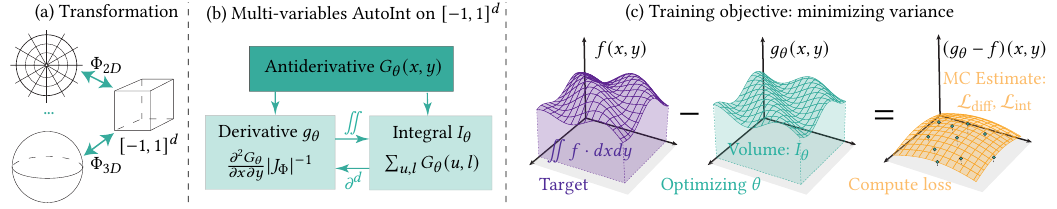}
    \vspace{-2em}
    \caption{
    Overview of our method. 
    (a) We first create a diffeomorphic transformation $\Phi$ that maps integration domain to a hyper-cube $[-1,1]^d$.
    (b) We generalize AutoInt~\cite{lindell2021autoint} to hyper-cube $[-1, 1]^d$ (Sec~\ref{sec:spatial-integral}).
    (c) During training, we directly minimize the variance of the estimator using Monte Carlo estimation (Sec~\ref{sec:loss}). 
    }
    \label{fig:method}
    \vspace{-1.em}
\end{figure*}

\section{Related Work}
We will focus on reviewing the most relevant papers in control variates and nueral integration techniques.
\vspace{-0.5em}
\paragraph{Control Variates.}
Control variates is an important variance reduction technique for Monte Carlo integration~\cite{glynn2002some,loh1995method,pajot2014globally}.
Prior works have applied control variates in many applications, including option pricing~\cite{ElFilaliEchChafiq2021AutomaticCV}, variational inference~\cite{Geffner2018UsingLE,Wan2019NeuralCV}, and Poisson image reconstruction~\cite{rousselle2016image}.
To establish a control variate, we need to find a function with a known analytical integration while correlating the integrand function well.
Most prior works usually construct the control variate heuristically~\cite{lafortune1994using,clarberg2008exploiting,kutz2017spectral}.
Such an approach can be difficult to generalize to complex integrands.
One way to circumvent such an issue is to make the control variates learnable and optimize the control variates function using samples from the integrand.
For example, \citet{salaun2022regression} proposed to use a polynomial-based estimator as control variate as the integration of the polynomial basis is easy to obtain.
Recently, \citet{muller2020ncv} proposed to use normalizing flow as the control variate function since normalizing flows are guaranteed to integrate into one.
Our method extends these works by expanding the choice of estimator family to a broader class of neural network architecture.
Most existing works apply CV on physics-based rendering.
We focus on applying CV to solving PDEs using Walk-on-sphere methods~\cite{WoS,VWos,WoSt}, which allows us to showcase the advantage of having a broader class of control variate function.

Existing works have attempted various techniques to reduce variances of the Walk-on-sphere algorithms, including caching~\cite{B-cache,sacache,bakbouk2023mean}, heuristic-based control variates~\cite{WoS, MCFluid}, and bidirectional formulations~\cite{qi2022bidirectional}.
These methods are orthogonal to our paper, which applies neural control variates method to reduce variance for Walk-on-Sphere algorithms.
In our experiment, we demonstrate that our method can be combined with existing variance reduction methods to reach better performance.
\vspace{-0.5em}
\paragraph{Neural Network Integration Methods.}
Deep learning has emerged as a dominant optimization tool for many applications, including numerical integration estimation. 
A prevalent strategy involves crafting specialized neural network architectures with analytical integration capabilities, similar in spirit to the Risch or Risch-Norman algorithm \cite{Risch1969ThePO,norman1977implementing}. 
For example, normalizing flows~\cite{tabak2013family,Dinh2016DensityEU,chen2018neural,dinh2014nice} is a family of network architectures that models an invertible mapping, which allows them to model probability distribution by integrating into one.
Other examples include \cite{petrosyan2020neural} and \cite{subr2021q}, which designed network architectures that can be integrated analytically.
These approaches usually result in a limited choice of network architectures, which might limit the expressivity of the approximator.
An alternative approach is to create computational graphs that can be integrated into a known network by taking derivatives.
For example, \cite{nsampi2023neural} leverages repeated differentiation to compute convolutions of a signal represented by a network.
In this work, we follow the paradigm proposed by AutoInt \cite{lindell2021autoint}, where we construct the integrand by taking network derivatives approximating the integration result.
This approach can allow a more flexible choice of network architectures, and it has been successfully applied to other applications such as learning continuous time point processes~\cite{zhou2023automatic}.
Unlike the Monte Carlo integration, a potential drawback to the AutoInt method is that it can create biased estimations.
In this work, we propose to combine the AutoInt method with neural control variate to create an unbiased estimator.


    

\section{Background}

In this section, we will establish necessary notations and mathematical background to facilitate the discussion of our method.
In particular, we will cover backgrounds in Monte Carlo integration, Control variates, and neural integration methods in line integration.

\paragraph{Monte Carlo Integration}
The main idea of Monte Carlo integration is to rewrite the integration into an expectation, which can be estimated via sampling. 
Assume we want to estimate the integration of a real-value function $f:\Re^d\to\Re$ over domain $\Omega$.
We first write it into an expectation over the domain $\Omega$:
\begin{align}
    \int_{x\in\Omega} f(x) dx
    = \int_{x\in\Omega} P_\Omega(x) \frac{f(x)}{P_\Omega(x)} dx
    = \mathbb{E}_{x \sim P_\Omega}\left[\frac{f(x)}{P_\Omega(x)}\right],
\end{align}
where $P_\Omega$ is a distribution over domain $\Omega$ from which we can both sample and evaluate likelihood.
This allows us to define the following estimator:
$
    \langle F_N \rangle = \frac{1}{N}\sum_{i=1}^N \frac{f(x_i)}{P_\Omega(x_i)}
$
,
where $x_i$'s are points sampled from $P_\Omega$ and $N$ denotes the number of samples.
Monte Carlo estimation is unbiased given that $P_\Omega(x) = 0$ only if $f(x) = 0$. 
However, it usually suffers from high variance, requiring a lot of samples and function evaluations to obtain an accurate result.
\paragraph{Control Variates.}
Control variates is a technique to reduce variance for Monte Carlo estimators.
The key idea is to construct a new integrand with lower variance and apply Monte Carlo estimation for the low-variance integrand only.
Suppose we know $G=\int_\Omega g(x)dx$ for some $G$ and $g$, then we have:
\begin{align}
\label{eq:spatial-cv}
    \int_\Omega f(x)dx = G + \int_\Omega f(x) - g(x) dx.
\end{align}
With this identity, we can derive a single-sample numerical estimator $\langle F_{cv} \rangle$ that is unbiased:
\begin{align}
    \langle F_{\text{cv}} \rangle = G + \frac{f(x_i) - g(x_i)}{P_\Omega(x_i)},\quad \text{where } x_i \sim P_\Omega.
\end{align}
As long as $G$ is the analytical integration result of $g$, the new estimator created after applying control variate is unbiased.
Note that the control variate estimator is running Monte Carlo integration on the new integrand $f - g$, instead of the original integrand $f$.
The key to a successful control variate is finding corresponding functions $G$ and $g$ that make $f - g$ to have less variance compared to the original integrand under the distribution $P_\Omega$.
Choosing an appropriate $g$ is challenging since it requires correlation with $f$ while having an analytical integral $G$.
Existing works either pick $g$ heuristically (\eg~$g=\cos x$ if $\cos x$ is a component of $f$), or use a limited family of parametric functions to approximate $f$ from data.
Our method circumvents this issue by learning a parametric model for the antiderivative of $g$, allowing us to use arbitrary neural network architecture for control variate.

\paragraph{Neural Integration.}
AutoInt~\cite{lindell2021autoint} proposes a way to estimate an integral using neural networks.
Suppose we want to estimate a line integral of the form $\int_L^U f(x) dx$, where $f$ is a real-value function and $L\leq U \in \Re$ denotes the lower and upper bound for integration.
AutoInt trains a neural network $G_\theta: \Re\to\Re$ to approximate the antiderivative of the integrand $f$ with learnable parameter $\theta$.
By the first fundamental theorem of calculus, we know that if for all $x$ in $[L,U]$, $\frac{\partial}{\partial x} G_\theta(x) = f(x)$, then:
\begin{align}
    \int_L^U f(x) dx = \int_L^U \frac{\partial}{\partial x}G_\theta(x) dx = G_\theta(U) - G_\theta(L).
\end{align}
To find the parameter $\theta$ that satisfies the constraint $\frac{\partial}{\partial x} G_\theta(x) = f(x) \forall x \in [L,U]$, AutoInt uses gradient-based optimizer to solve the following optimization problem:
\begin{align}
    \theta^* = \argmin_{\theta} \Ee_{x\in \uU[L,U]} \left[ \paren{f(x) - \frac{\partial}{\partial x}G_\theta(x)}^2 \right],
\end{align}
where $\uU[L,U]$ is uniformly distributed over interval $[L, U]$ and the derivative $\frac{\partial}{\partial x}G_\theta(x)$ is obtained via the automatic differentiation framework.
Once the network is trained, we can use optimized parameters $\theta^*$ to approximate the integration results of $\int_L^U f(x)dx$.
This idea can be extended to multi-variables integration~\citep{maitre2023multi} by taking multiple derivatives, which we will leverage in the following section to construct integrations for different spatial domains, such as disk and sphere.

Neural integration methods have several advantages.
First, one can use arbitrary neural network architecture with this method.
This allows users to leverage the latest and greatest network architectures, such as SIREN~\cite{sitzmann2020implicit} and instant-NGP~\cite{mueller2022instant}, potentially leading to better performance.
Second, neural integration can approximate a family of integrals.
The abovementioned example can approximate integration of the form $\int_l^u f(x)dx$ for all pairs of $l \leq u$ such that $L \leq l \leq u \leq U$.
AutoInt~\cite{lindell2021autoint} also show that one can modulate the network $G_\theta$ to approximate a family of different integrand.
However, it's difficult to guarantee that the network $\frac{\partial}{\partial x}G_\theta$ can approximate the integrand exactly as the loss is difficult to be optimized to exactly zero.
In this paper, we can circumvent such an issue as we use AutoInt inside control variates.
Our method can both enjoy the advantages brought by neural integration methods while maintaining the guarantee provided by monte carlo integration methods.

\section{Method}
In this section, we will demonstrate how to use neural integration method to create control variates functions from arbitrary neural network architectures.
We will first demonstrate how to construct networks with known analytical spatial integrals (Sec~\ref{sec:spatial-integral}).
We then show how to create a numerically stable unbiased estimator using these networks as control variates (Section~\ref{sec:ncv}) as well as a numerically stable training objective aiming to minimize the variance of the estimator (Sec~\ref{sec:loss}).
Finally, we will discuss how to extend this formulation to multiple domains (Sec~\ref{sec:multi-domains}).
\vspace{-.5em}
\subsection{Neural Spatial Integration}\label{sec:spatial-integral}
Computer graphics applications, such as rendering and solving PDEs, usually require integrating over spatial domains such as sphere and circles.
To make neural integration methods applicable to these applications, we need to adapt them to integrate over various spatial domains by applying change of variables.

Let's assume the integration domain $\Omega \subset \Re^d$ is parameterized by an invertible function $\Phi$ mapping from a hypercube $U = [-1,1]^d$ to $\Omega$, \ie~$\Phi(U) = \Omega$.
For any neural network $G_\theta:U\to\Re$, we can apply the first fundamental theorem of calculus to obtain the following identity~\cite{maitre2023multi}:
\vspace{-.5em}
\begin{align}
    \label{eq:multi-var}  
    \int_U \frac{\partial^d}{\partial \uu} G_\theta(\uu)d\uu = 
    \underbrace{
    \sum_{u_1\in\{-1, 1\}}\dots \sum_{u_d\in\{-1, 1\}} G_\theta(\uu) \prod_{i=1}^du_i}_{
    \text{Defined as } I_\theta
    },
\end{align}
\vspace{-.5em}
where $\uu=[u_1, \dots, u_d]$ and $\frac{\partial^d}{\partial \uu} G_\theta(\uu)$ denotes partial derivative with respect to all dimension of vector $\uu$: $\frac{\partial^d}{\partial \uu_1\dots \partial \uu_d} G_\theta(\uu)$.
Note that we can obtain both the computation graph for the integrand $\frac{\partial^dG_\theta}{\partial \uu}$ and the right-hand-side $I_\theta$ using existing deep learning frameworks with automatic differentiation.
To extend this idea to integrating over $\Omega$, we need to apply the change of variable:
\vspace{-.5em}
\begin{align}
    \label{eq:change-var}
    I_\theta = \int_U \frac{\partial^d}{\partial \uu} G_\theta(\uu)d\uu
    = \int_\Omega \frac{\partial^d}{\partial \uu} G_\theta(\uu)\left|J_\Phi(\uu)\right|^{-1}d\xx,
\end{align}
where $\xx$ are coordinate in domain $\Omega$, $\uu = \Phi^{-1}(\xx)$, and $J_\Phi \in \Re^{d\times d}$ is the Jacobian matrix of function $\Phi$.
Since the integrand from the right-hand-side can be obtained through automatic differentiation, we now obtain a optimizable neural network with known integral in domain $\Omega$.
This identity is true regardless of the neural network architecture.
This opens up a rich class of learnable parametric functions useful for control variates.
Table~\ref{tab:integral} shows $\Phi$ and $|J_{\Phi|}$ in three integration domains: 2D circle, 2D disk, and 3D sphere.
\begin{table}[t]
  \centering
  \small
  \caption{Transformation and Jacobian for variable spatial domains. Note that we assume input domain $U=[-1,1]^d$.
  }
  \vspace{-1em}
    \begin{tabular}{lcc}
    \toprule
    Domain $\Omega$ & $\Phi:U\to\Omega$ & $|J_\Phi|$\\
    \midrule
    2D Circle       & $\theta \mapsto (\cos(\pi\theta), \sin(\pi\theta))$ & 1\\
    2D Disk         & $(r, \theta) \mapsto \frac{r + 1}{2}\cdot (\cos(\pi\theta), \sin(\pi\theta))$ & $\frac{r + 1}{2}$ \\
    3D Sphere       & $\begin{bmatrix}\theta \\ \phi\end{bmatrix} \mapsto \begin{bmatrix}\sin(\pi(\phi+1)/2)\cos(\pi(\theta+1))\\ \sin(\pi(\phi+1)/2)\sin(\pi(\theta+1)) \\ \cos(\pi(\phi+1)/2)\end{bmatrix}$ &   $\frac{1}{2}\sin\paren{\frac{\pi(\phi+1)}{2}}$\\
    \bottomrule
    \vspace{-2.8em}
    \end{tabular}%
  \label{tab:integral}%
\end{table}%

\vspace{-.5em}
\subsection{Control Variates Estimator}\label{sec:ncv}
Equation~\ref{eq:change-var} now allows us to construct neural networks with analytical integral for various spatial domains such as 2D circles, 2D disks, 3D spheres, and more.
These neural networks can be used for neural control variates, substituting Equation~\ref{eq:change-var} into Equation~\ref{eq:spatial-cv}:
\begin{align}
    \int_{\Omega} f(\xx) d\xx
    = I_\theta + \int_{\Omega} f(\xx) - 
    \frac{\partial^d }{\partial \uu}G_\theta(\uu)\left|J_\Phi(\uu)\right|^{-1}d\xx,
\end{align}
where $\uu=\Phi^{-1}(\xx)$.
Now we can create a single-sample control variates estimator $\langle F_{\text{ncv}}(\theta) \rangle$ to approximate the spatial integration:
\begin{align}\label{eq:ncv-est}
    \langle F_{\text{ncv}}(\theta) \rangle = I_\theta + \frac{f(\xx_i)}{P_{\Omega}(\xx_i)}
         - 
        \frac{\frac{\partial^d }{\partial \uu}G_\theta(\uu_i)
        }{\left|J_\Phi(\uu_i)\right|P_{\Omega}(\xx_i)}
        ,
\end{align}
where $\xx_i \sim P_\Omega$ are independent samples from a distribution on the domain $\Omega$, $P_\Omega(\xx_i)$ is the probability density of $\xx_i$ , and $\uu_i=\Phi^{-1}(\xx_i)$.
While $\langle F_{\text{ncv}}(\theta) \rangle$ is unbiased, it is numerically unstable when $|J_\Phi|$ takes a very small value.
For example, when integrating a 2D disk, estimator $\langle F_{\text{ncv}}(\theta) \rangle$ is unstable when $\frac{r+1}{2}$ is near $0$.
To tackle this issue, we first change the transformation function $\Phi$ to map $U$ to a numerically stable domain $\Omega_\epsilon$. 
For the case of 2D disk, we replace $\Phi$ with $\Phi_\epsilon = \Phi \circ T_\epsilon$, where $T_\epsilon(r, \theta) = (r(2 - \epsilon)/ 2 + \epsilon, \theta)$, with a small number $\epsilon\in\Re$.
The intuition behind $T_\epsilon$ is to map the domain $U$ to a place where $\frac{r+1}{2}$ is not near zero.
We can use such a transformation $\Phi_\epsilon$ to create an unbiased estimator as following:
\vspace{-.5em}
\begin{align}
\label{eq:est-ncv-eps}
 \langle F_{\text{ncv}}(\theta) \rangle = I_\theta + \frac{f(\xx_i)}{P_{\Omega}(\xx_i)}
    - 
 \mathds{1}_U(\uu_i)\frac{\frac{\partial^d }{\partial \uu}G_\theta(\uu_i)}{\left|J_{\Phi_\epsilon}(\uu_i)\right|P_{\Omega}(\xx_i)}
,
\end{align}
where $\xx_i\sim P_\Omega$, $\uu_i = \Phi_\epsilon(\xx)$, and $\mathds{1}_U(\uu_i)$ is an indicator function that equals to $1$ if $\uu_i \in U$ and returns $0$ otherwise.

If $\theta$ is not chosen intelligently, $\langle F_{\text{ncv}}(\theta) \rangle $ can have higher variance than directly estimating the original integrand $f$.
We will show in the upcoming section how to minimize the variance of such an estimator using deep learning tools.
\vspace{-.5em}
\subsection{Training Objectives: Minimizing Variance}\label{sec:loss}
Our networks can be trained with different loss functions, and one should choose the loss function that works the best depending on the specific application.
In this section, we will use the estimator's variance as an example to demonstrate how to adapt a control variate loss function to train our model.
Following ~\citet{muller2020ncv}, the variance of the estimator $\Ve\left[\langle F_{\text{ncv}}(\theta) \rangle\right]$ in Eq.~\ref{eq:est-ncv-eps} is:
\vspace{-.5em}
\begin{align*}
\int_\Omega\frac{\paren{
        f(\xx) - 
        \mathds{1}_U(\uu)\frac{\partial^d G_\theta(\uu)}{\partial \uu}|J_\Phi(\uu)|^{-1}
    }^2}{P_{\Omega}(\xx)}
    d\xx
-
\paren{I_\theta - \int_\Omega f(\xx)d\xx}^2,
\end{align*}
where $\uu = \Phi^{-1}(\xx)$. 
Directly using this variance as a loss function is infeasible since we do not have analytical solutions for the term $\int_\Omega f(\xx)d\xx$.
Since the gradient-based optimizer only requires gradient estimate to be unbiased, one way to circumvent this issue is to create a pseudo loss whose gradient is an unbiased estimator for the gradient of $\Ve\left[\langle F_{\text{ncv}}(\theta) \rangle\right]$.
First, we define the following losses:
\begin{align}
\lL_{\text{int}}(\theta, \Omega) &= \Ee_{\xx\sim \uU(\Omega)}\left[  (f(\xx)|\Omega| - I_\theta)^2\right], \\
\lL_{\text{diff}}(\theta, \Omega) &= \Ee_{\xx\sim P_\Omega}\left[  
\frac{\paren{
        f(\xx) - 
        \mathds{1}_U(\uu)\frac{\partial^d G_\theta(\uu)}{\partial \uu}|J_\Phi(\uu)|^{-1}
    }^2}{P_{\Omega}(\xx)^2}
\right], 
\end{align}
where $\uU(\Omega)$ denotes uniform sampling of domain $\Omega$.
One can verify that $\nabla_\theta \Ve\left[\langle F_{\text{ncv}}(\theta) \rangle\right] = \nabla_\theta \lL_{\text{diff}}(\theta, \Omega) - \nabla_\theta \lL_{\text{int}}(\theta, \Omega)$ (See supplementary).
Note that $\lL_{\text{diff}}$ can numerically unstable when $|J_\Phi(\uu)|$ is very small.
Since we are using $\Phi_\epsilon$ in Equation~\ref{eq:est-ncv-eps}, we can see that $\nabla_\theta \lL_{\text{diff}} = 0$ in the region when $|J_\Phi(\uu)|$ is very small.
As a result, we discard those samples during training.
Note that similar techniques can be applied to other types of control variates losses. 

\vspace{-.5em}
\subsection{Modeling a Family of Integrals}\label{sec:multi-domains}
So far we've focused on applying our method to a single integration $\int_\Omega f(\xx)d\xx$ over a single domain $\Omega$.
In many computer graphics applications, we need to perform multiple spatial integrals, each of which will be using a slightly different domain and integrand. 
Adapting to such applications, we need to apply CV to solve a family of integrations in the form of
$
    \int_{\Omega(\cc)} f(\xx, \cc) d\xx,
$
where $\cc\in\Re^h$ is a latent vector parameterizing the integral domain, $\Omega(\cc) \subset \Re^d$ are a set of domains changing depending on $\cc$, and $f$ is the integrand.

One way to circumvent this challenge is to learn coefficients for one CV that minimize the variance of all estimators as proposed by \citet{hua2023revisiting}.
In our paper, we choose an alternative way, training a conditional neural network that can predict CV functions for the whole family of integrals.

To achieve this, we first assume there exists a family of parameterization functions for this family of domains $\Phi:\Re^d\times\Re^h\to\Omega$, where each function $\Phi(\cdot, \cc)$ is differentiable and invertible conditional on $\cc$, and $\Phi(U, \cc) = \Omega(\cc)$.
Now we can extend our network $G_\theta$ to take not only the integration variable $\xx$ but also the conditioning latent vector $\cc$. 
We will also extend the loss function to optimize through different latent $\cc$:
$
    \lL_{\text{multi}}(\theta) =\frac{1}{N}\sum_{i=1}^N \lL_{\text{diff}}(\theta,\Omega(\cc_i)) - \lL_{\text{int}}(\theta,\Omega(\cc_i)) 
$.
The same principles described in previous sections will still apply.

\section{Results}
In this section, we will provide a proof of concept showing that our method can be applied to reduce the variance of Walk-on-Spheres algorithms (WoS)~\cite{WoS}.
We will first demonstrate that our method creates unbiased estimators in different integration domains while using different neural network architectures (Sec~\ref{sec:result-unbiased}).
We then evaluate our method's effectiveness in solving 2D Poisson and 3D Laplace equations (Sec~\ref{sec:equal-sample}, Sec~\ref{sec:wall-time}).

\subsection{Unbiased Estimator with Arbitrary Network}\label{sec:result-unbiased}
\begin{figure}[t]
    \centering
    \begin{tabular}{c@{}c@{}c}
         2D Circle &
         2D Disk &
         3D Sphere 
         \\
         \includegraphics[width=0.34\linewidth,height=0.32\linewidth]{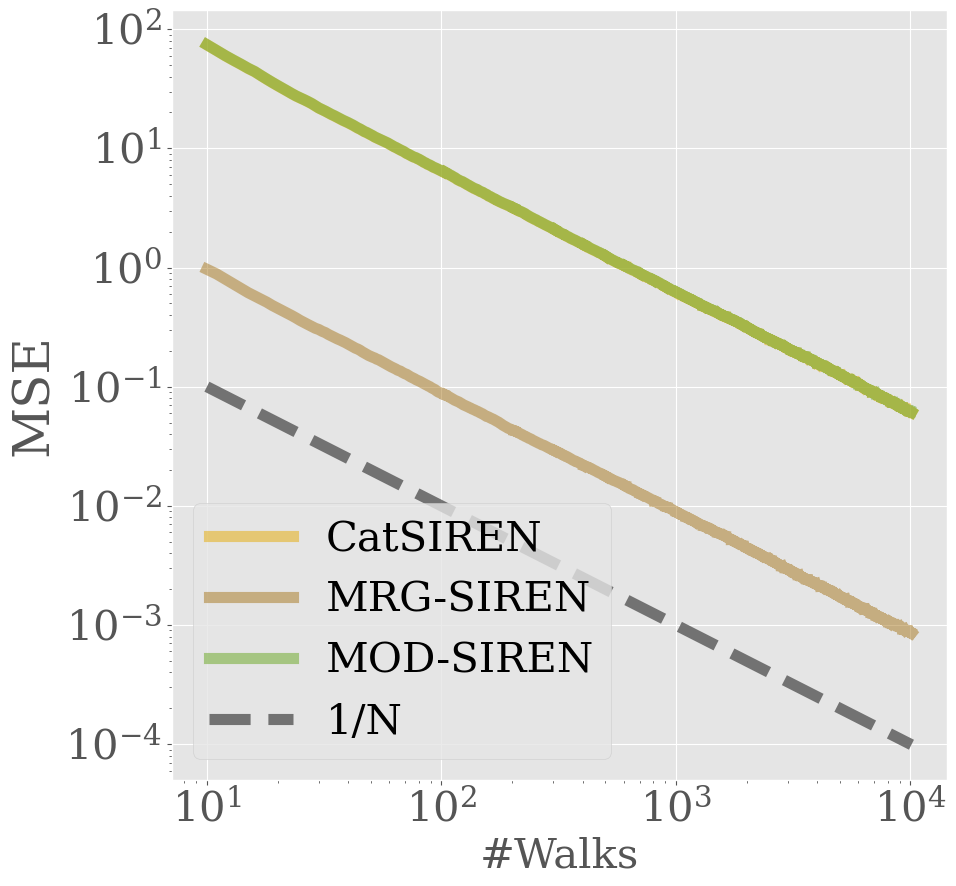} &
        \includegraphics[width=0.31\linewidth,height=0.32\linewidth]{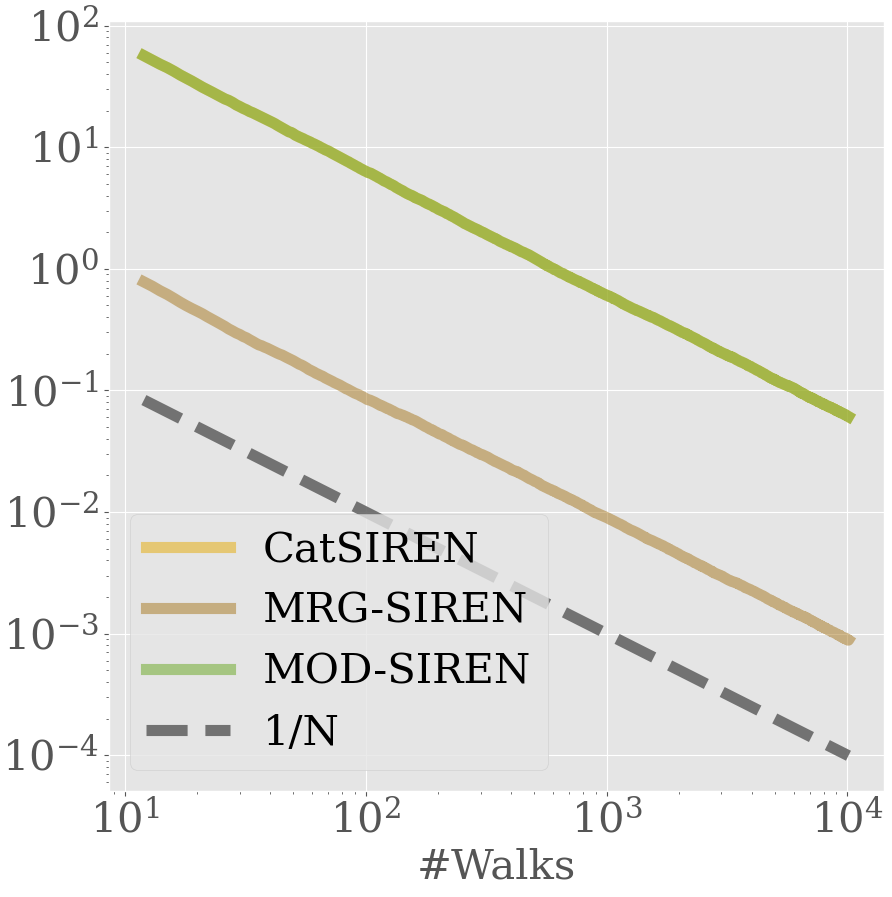} &
        \includegraphics[width=0.31\linewidth,height=0.32\linewidth]{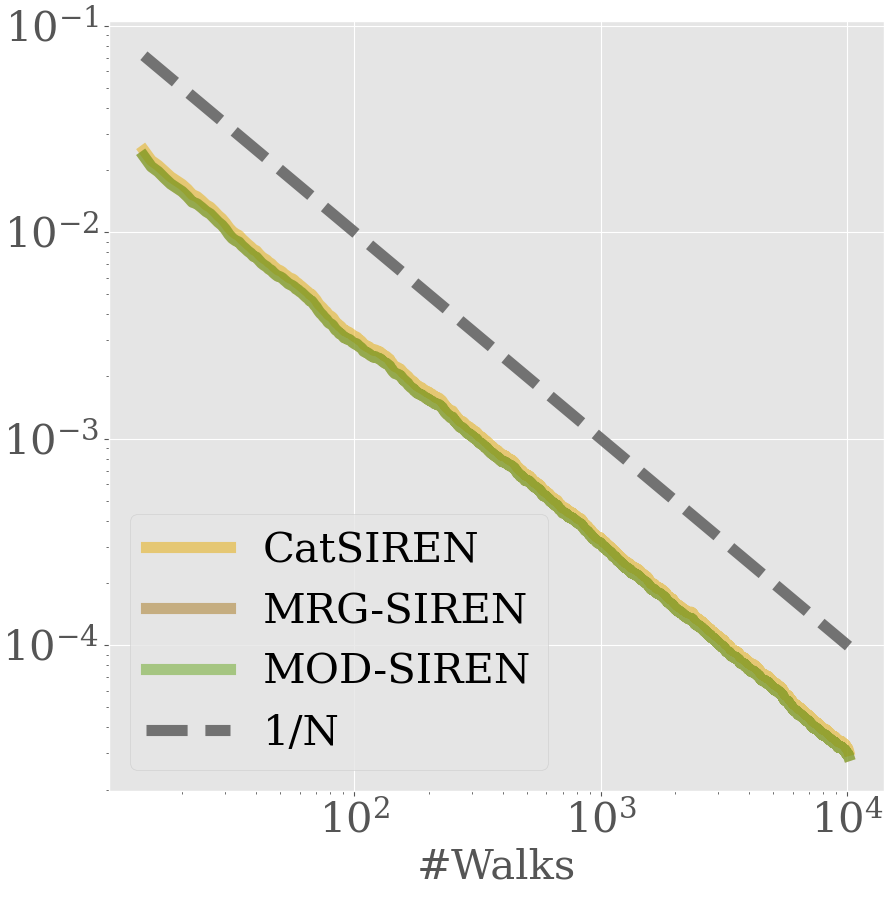} 
    \end{tabular}
    \vspace{-1.5em}
    \caption{Convergence curve of our CV estimator using various randomly initialized networks. 
    This suggests that our method can produce unbiased control variates estimators from arbitrary network architectures.
    }
    \vspace{-2.em}
    \label{fig:rand-init}
\end{figure}

In this section, we want to show that our method indeed creates an unbiased estimator regardless of neural network architectures or the integration domain.
We will test our method on three types of integration domains mentioned in Table~\ref{tab:integral}: 2D circle, 2D disk, and 3D spheres.
We will test the following neural network architectures:

\noindent
\paragraph{CatSIREN} 
\citet{sitzmann2020implicit} proposed SIREN, which uses periodic activation to create an expressive neural network capable of approximating high-frequency functions. 
We make this network architecture capable of taking conditioning, we simply concatenate the condition vector $\cc$ with the integration variable $\uu$:
\begin{align*}
\phi_i(\zz) = \sin(\WW_i\zz + \bb_i),
\ 
G_\theta(\xx, \cc) = \WW_n(\phi_{n-1} \circ \dots \circ \phi_0)([\uu, \cc]) + \bb_n,
\end{align*}
where $\theta=\{\WW_i, \bb_i\}_i$ are the trainable  parameters.

\noindent
\paragraph{ModSIREN}
\citet{mehta2021modulated} proposed a way to condition SIREN network more expressively using a parallel ReLU-based network to produce the frequency of SIREN's periodic activate:
\begin{align*}
\hh_0 &= \max(\WW_0'\cc + \bb_0', 0), &\hh_{i+1} &=\max(\WW_{i+1}'[\cc, \hh_i] + \bb_{i+1}', 0) \\
\phi_i(\zz) &= \sin(\WW_i\zz \odot \hh_i + \bb_i), &G_\theta(\xx, \cc) &= \WW_n(\phi_{n-1}\circ \dots \circ \phi_0)(\xx) + \bb_n,
\end{align*}
where $\theta=\{\WW_i, \WW_i', \bb_i, \bb_i'\}_i$ are trainable parameters.

\paragraph{MGC-SIREN}
If the conditioning latent code is low dimensional, such as $\Re^2$ or $\Re^3$ in our applications, we could borrow the idea from instant-NGP~\cite{mueller2022instant}, in which we can modulate the network using a spatially varying feature grid.
Basically, we use $\cc$ to extract a latent feature $\zz$ from a multi-resolution grid.
We then use the extracted feature $\zz$ as the conditional feature in \verb|CatSIREN| architecture.
The trainable parameters include both the feature grid and the network parameters.

In this experiment, we initialize a ground truth field $u:\Re^d\to\Re$ in a cube $[0, 1]^d$ and deploy these network to produce estimator for following family of integrals:
$
    F(x) = \int_{\Omega(x)} u(x+y) dy,
$
where $x\in[0,1]^2$ is a coordinate in the domain of interest and $\Omega(x)$ denotes the integration domain centered at $x$.
For 2D circle and 2D sphere, we set $\Omega(x) = \{y\ | \norm{x-y} = d(x)\}$, where $d(x)$ is the distance to the nearest point of the boundary $[0, 1]^2$.
For 2D ball, we set $\Omega(x) = \{y\ | \norm{x-y} \leq d(x)\}$.

We randomly initialize the abovementioned three network architectures and present the MSE between the reference solution and their corresponding CV estimators (Eq~\ref{eq:est-ncv-eps}) to ground truth.
Figure~\ref{fig:rand-init} shows how mean square errors decay with the number of samples.
Our method produces unbiased estimators for even randomly initialized neural networks with different architectures.
We will stick with \verb|MGC-SIREN| architecture for the rest of the section.
\subsection{Equal Sample Comparisons}\label{sec:equal-sample}

In this section, we will focus on providing equal sample analysis comparing our methods with prior arts on the task of reducing the variance of WoS algorithms in solving 2D Poisson and 3D Laplace equations.
Equal sample comparisons are useful because they are less confounded with implementation and hardware details. 
For example, engineering techniques such as customized CUDA kernels can drastically affect the compute time on our estimator, but they won't affect the result of equal sample analysis as much.

\paragraph{Baselines.}
We compare our methods with WoS without control variates\footnote{The original control variates method mentioned at \cite{WoS} can be viewed as a special case for the \citet{salaun2022regression} with degree 1. As a result, we do not include it as one of the baselines.} and two other learning-based control variates baselines.
The first baseline is \citet{muller2020ncv} (\verb|NF|), which uses normalizing flows to parameterize the control variates function.
The second baseline is \citet{salaun2022regression} (\verb|POLY|), which parameterize using a weighted sum of polynomial basis.
We follow the original papers on hyperparameters, such as the degrees of the polynomial-basis and positional encoding. 
Specifically, we use a polynomial order of 2 following the suggestions of Fig 6 at \citet{salaun2022regression}.
For fair comparison, we use the same loss functions and optimizer to train the neural CV baselines. 
We apply multi-resolution grids mentioned in Section~\ref{sec:result-unbiased} to allow \verb|POLY| baseline to perform CV at arbitrary locations within the PDE domain.
Specifically, the multi-resolution grid takes as input the position of the walk (\ie center of the ball/sphere), and the interpolation of the grid will provide a vector that's later decoded to the polynomial coefficients of the integrand at the grid location.
We use a single linear layer to map the interpolated latent code to the set of coefficients.

\paragraph{Training.}
We create a data cache of size 16384 and use samples from about $10$ percent of total inference walks for training. 
All network is trained for $25000$ optimization steps using Adam\cite{kingma2014adam} optimizer with a learning rate of $10^{-4}$ and batch size $1024$. 
For every training iteration, we update $1024$ data cache locations.
For each updated point, we will create a label that accumulates $32$ walks to store in the data cache.
All experiments are conducted with a single RTX 2080 Ti GPU.

\begin{figure}[t]
    \centering
    \includegraphics[width=0.48\textwidth]{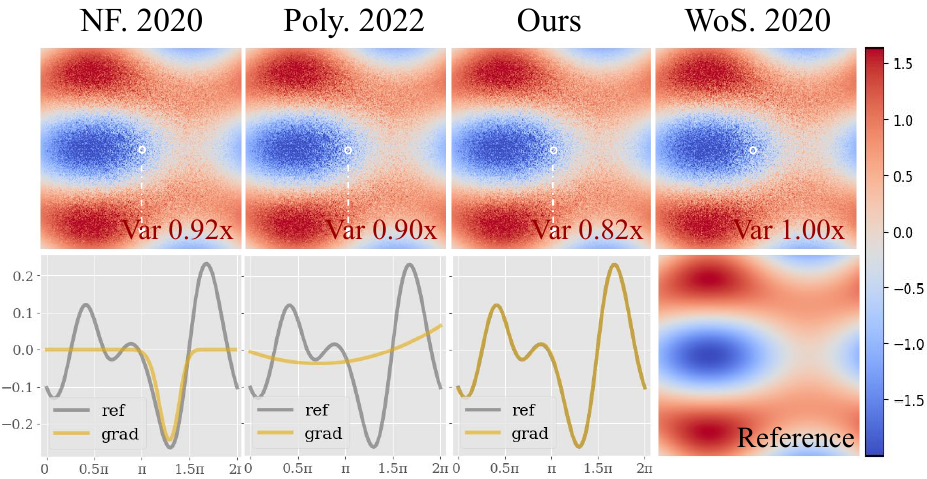}
    \vspace{-2em}
    \caption{
    Top: Equal sample comparison of solving 2D Poisson equations using control variates integrating over a 2D circle. Bottom: Plotting of Network prediction and integration reference.
    Our network can fit the integrand tightly.
    This leads to an estimator that produces the lowest variance.
    }
    \vspace{-1.5em}
    \label{fig:2d-circle}
\end{figure}

\vspace{-.5em}
\subsubsection{Solving 2D Poisson Equation}\label{sec:result-poisson}
We now apply our techniques to reduce variance on a Poisson equation over the domain $S\subset \Re^2$:
\begin{align}
    \Delta u = f  \text{ on } S,\quad u = g \text{ on } \partial S,
\end{align}
where $g:\Re^2\to\Re$ is the boundary function, and $f:\Re^2\to\Re$ is the forcing function.
This equation can be solved by the following integral equation~\cite{WoS}:
\begin{align*}
    u(x) = \int_{\partial B_{d(x)}(x)} \frac{u(y)}{|\partial B_{d(x)}(x)|}dy +\int_{B_{d(x)}(x)}f(y)G(x, y)dy,
\end{align*}
where $d(x) = \min_{y\in\partial \Omega}\norm{x-y}$ denotes the distance to the boundary and $B_r(c)=\{y | |y-c| \leq r\}$ is the ball centered at $c$ with radius $r$.
\cite{WoS} further derives a Monte Carlo estimator $\hat{u}(x)$ for the Poisson equation:
\begin{align}\label{eq:wos-3dpos}
    \hat{u}(x) = \begin{cases}
    g(x)\ \ &\text{if } d(x) < \epsilon \\
    \hat{u}(x') - |B_{d(x)}(x)|f(y)G(x, y)\ \ &\text{otherwise}
    \end{cases},
\end{align}
where $x'\sim \uU(\partial B_{d(x_k)}(x_k))$, $y \sim \uU(B_{d(x_k)}(x_k))$ are samples from the boundary and the interior of the 2D disk, and $G$ denotes the 2D disk's green's function.
These are two integrals that our method can be applied to: one that integrates over the circle (\ie~$\partial B$) and one that integrates inside the disk (\ie~$B$).

\paragraph{Apply Control Variates to 2D Circle $\partial B(x)$}
We first discuss how to apply our method to reduce variance of the integral $\int_{\partial B}\frac{u(y)}{|\partial B|}dy$. 
We first instantiate a neural network $G_\theta(t, x)$ that takes a sample's polar angle $t$ (normalized to $[-1, 1]$) and the 2D coordinate of the center of the disk $\partial B$.
$G_\theta$ outputs the anti-derivative of $\frac{u(y)}{|\partial B|}$.
Applying our methods, we can arrive at the following estimator:
\begin{align}
    \hat{u}_{\text{rec}}(x) = \begin{cases}
    g(x)+D(x,x') &\text{if } d(x_k) < \epsilon \\
    \hat{u}_{\text{rec}}(x') + S(x,y) + D(x,x')
    &\text{otherwise}
    \end{cases},
\end{align}
where $t$ is the polar angle of vector $x'-x$, $S(x,y)$ is the single sample estimator of the forcing contribution $|B(x)|f(y)G(x,y)$, and $ D(x,x') = I_\theta(x) - 2\pi \frac{\partial}{\partial t}G_\theta(t,x)$ captures the contribution of our control variates estimator.
We follow the training setup described in \ref{sec:equal-sample} with $\lL_{\text{diff}}$ loss that uses noise labels produced by $\hat{u}(x)$ to show our network can match the integrated closely.
The results are presented in Figure~\ref{fig:2d-circle}.

The POLY baseline fails to produce accurate results because the high-frequency integration pattern is hard to train for polynomials with low orders.
Similarly, this error also manifests in the NF baseline, even after we conducted a hyperparameter search to identify the best setting. 
The second row of Figure~\ref{fig:2d-circle} shows that both NF and POLY baseline tend to produce overly smoothed control variate functions, which makes the difference between the actual and predicted integrands still a high-variance function.

On the contrary, our method can better approximate the line integral on the circle $\partial B$, resulting in a smaller variance.

\begin{figure}[t]
    \centering
    \includegraphics[width=0.5\textwidth]{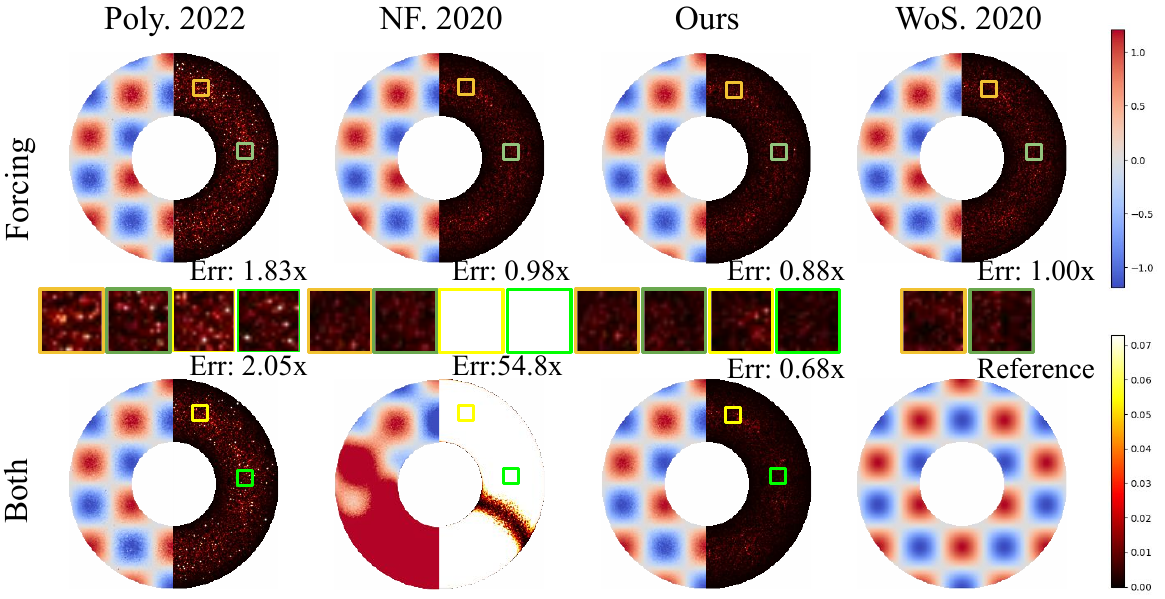}
    \vspace{-2em}
    \caption{
    Equal Sample Comparison of applying control variates (CV) when solving Poisson equations within a coin domain. 
    Top: apply CV on forcing integral; Bottom: apply CV on both forcing and recursive integral. In both settings, our method achieves the lowest error. 
    }
    \vspace{-1.5em}
    \label{fig:2d-disk}
\end{figure}

\vspace{-.5em}
\paragraph{Apply Control Variates to 2D Disk $B(x)$}
Our techniques can also be applied to reducing the variance of the family of integral over 2D disks: $\int_{B_{d(x)}(x)} f(y)G(x,y)dy$. 
Specifically, our antiderivative network $G'_\theta$ to take a normalized polar coordinate $p\in[-1,1]^2$ and the center of the circle $x\in\Re^2$.
Applying our method, we can construct the following single sample numerically stable control variates estimator for the forcing contribution:
\begin{align}
    S_{\text{cv}}(x,y)  = I'_\theta(x) + |B(x)|f(x)G(x,y) - \frac{2\mathds{1}(r)}{r + 1}\frac{\partial^2}{\partial p}G'_\theta(p,y),
\end{align}
where $r=\norm{y-x}$, $\mathds{1}$ is an indicator guarding the numerical stability, and $p$ is the polar coordinate of vector $y-x$.
Putting this estimator inside the WoS estimator gives us an unbiased estimator:
\vspace{-.5em}
\begin{align}
    \hat{u}_{\text{frc}}(x) = \begin{cases}
    g(x)&\text{if } d(x_k) < \epsilon \\
    \hat{u}_{\text{frc}}(x') + S_{\text{cv}}(x,y)
    &\text{otherwise}
    \end{cases}.
\end{align}
These two control variates techniques can be combined to account for both the variance from sampling sourcing contribution as well as the variance from recursive integration:
\begin{align}
    \hat{u}_{\text{both}}(x) = \begin{cases}
    g(x)+D(x,x')&\text{if } d(x_k) < \epsilon \\
    \hat{u}_{\text{both}}(x') + S_{\text{cv}}(x,y) + D(x, x')
    &\text{otherwise}
    \end{cases}.
\end{align}
Similarly, we use the same setup as in \ref{sec:equal-sample} and optimize the network using variance-reduction loss.
We present the results of these two estimators in Figure~\ref{fig:2d-disk}.
Our method consistently outperforms baselines, reaching the lowest error with the same number of samples.

\begin{figure*}
    \centering
    \vspace{-.5em}
    \includegraphics[width=0.99\linewidth]{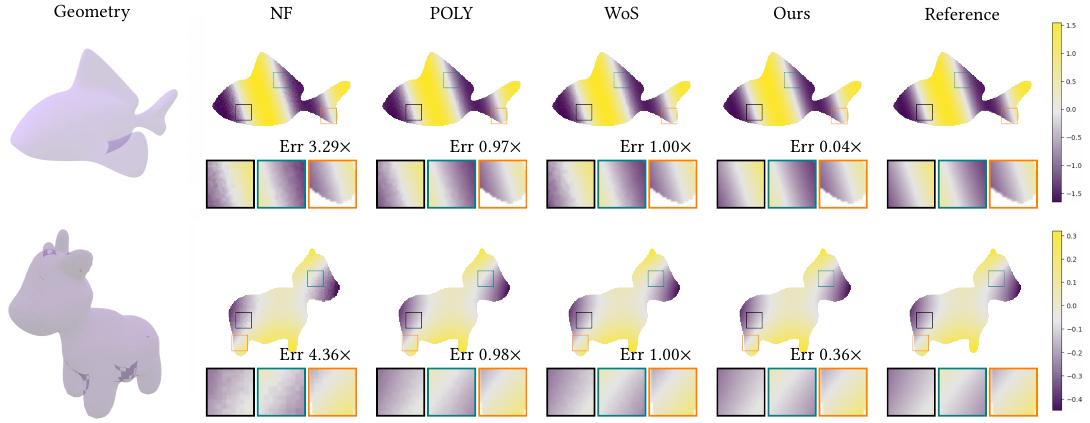}
    \vspace{-1em}
    \caption{Equal Sample Comparison. We visualize a 2D slice of the solution to solving Laplace Equations within the Blub shape domain. Our method produces less noisy results and achieves lower error than baseline methods.}
    \label{fig:3dlaplace}
    \vspace{-.5em}
\end{figure*}
\subsubsection{Solving 3D Laplace Equation}\label{sec:result-laplace}
In this section, we will demonstrate that our proposed method can also be applied to spherical integration.
Specifically, we will apply our method to solve 3D Laplace Equation using WoS methods:
\vspace{-.5em}
\begin{align}
    \Delta u = 0  \text{ on } S,\quad u=g \text{ on } \partial S,
\end{align}
where $S$ is the domain where we would like to solve the Equation equation and $g$ is the boundary condition.
Similar to the Poisson equation, the Laplace equations can be solved by the estimator in Eq~\ref{eq:wos-3dpos}, after replacing $B$ and $\partial B$ with their 3D counterpart and setting the forcing function to be zero.

Applying our framework, we will train a neural network $G_\theta(s, x)$, where $s\in\Re^2$ is a spherical coordinate normalized to $[-1, 1]^2$ and $x\in\Re^3$ is the conditioning which modulates the integration domain $\partial B_{d(x)}(x)$.
Applying our method, we can construct the following neural control variates estimator:
\vspace{-.5em}
\begin{align}
    \hat{u}_{\text{cv}}(x) &= \begin{cases}
    g(x)+D(x,x')\quad&\text{if } d(x) < \epsilon \\
    \hat{u}_{\text{cv}}(x') + D(x, x') &\text{otherwise}
    \end{cases}, \\
    D(x, x') &= I_\theta(x) - \frac{\partial^2}{\partial s} G_\theta(s, x)\paren{\sin \paren{\frac{\phi+1}{2}}}^{-1},
\end{align}
where $x'$ is a sample on the 3D sphere $\partial B(x)$, $s$ is the spherical coordinate of $x'-x$, and $\phi$ is the polar angle of the vector $\frac{x'-x}{\norm{x'-x}}$.

\setlength{\columnsep}{1em}
\setlength{\intextsep}{0em}
\begin{wrapfigure}{r}{0.5\linewidth}
\centering
\includegraphics[width=\linewidth]{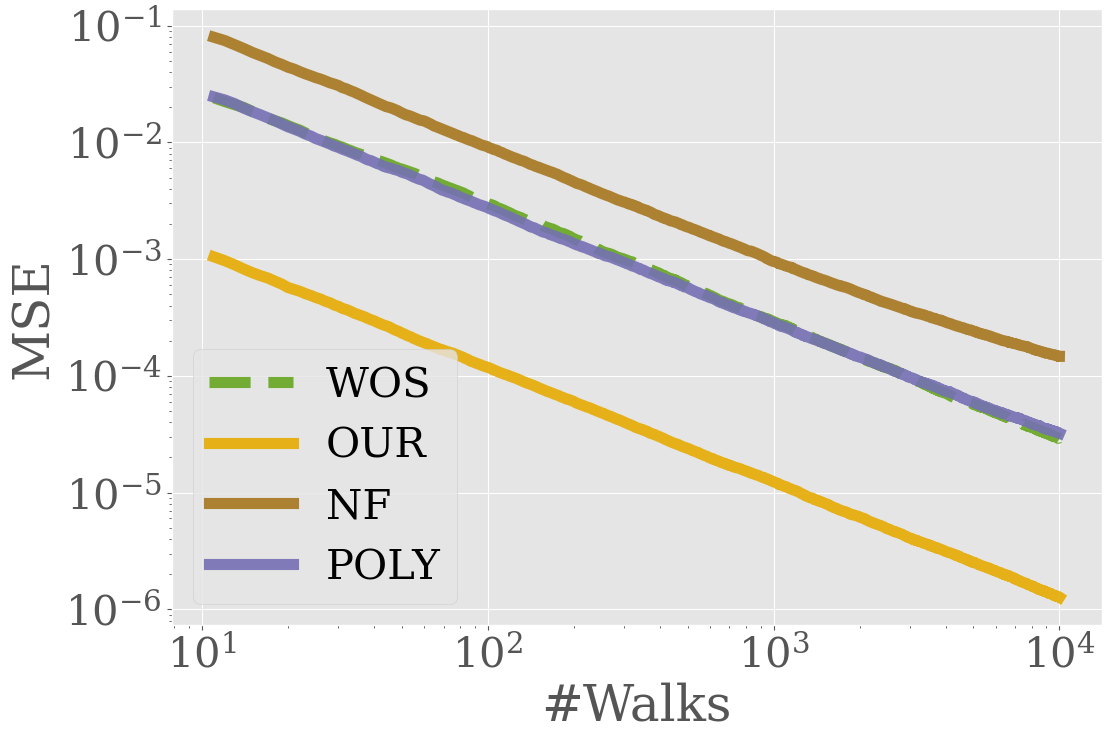}
\vspace{-2.5em}
\caption{
Our method is unbiased and achieves the lowest mean square errors.}
\vspace{-.5em}
\label{fig:convergence}
\end{wrapfigure}
Similar to the previous section, we obtain $\theta$ by optimizing the training objective mentioned in Section~\ref{sec:multi-domains} using Adam optimizer.
The training data is obtained using WoS estimator (Eq~\ref{eq:wos-3dpos} with $f=0$), which returns a noisy estimate of the integrand of interest.
We present the convergence curve of our method in comparison with the baselines in Figure~\ref{fig:convergence}.
We can see that under the same number of samples, our method achieves much lower MSE compared to the baseline.
We present the qualitative result in Figure~\ref{fig:3dlaplace} with Spot\cite{spot} and Blub\cite{blub}shapes and visualize a 2D slice of the solution.
All results are obtained under the same number of evaluation steps.
We can see that our estimator leads to a less noisy field compared to baselines.
\begin{figure}[t]
    \centering
    \includegraphics[width=0.45\textwidth]{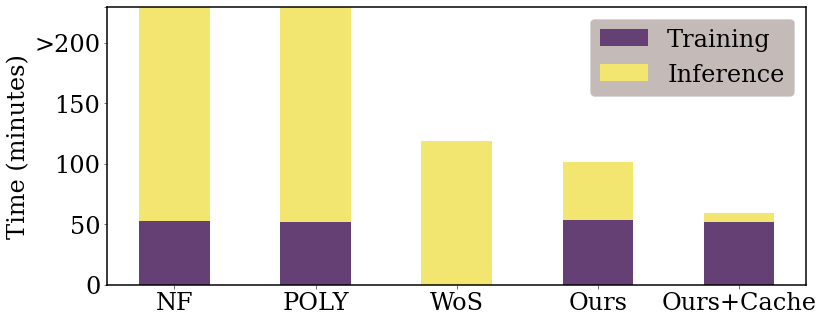}
    \vspace{-1em}
    \caption{We report the time needed for each method to create a solution in 1024x1024 resolution with <3e-5 MSE for the Spot shape. Note that within the plotted time range, NF and POLY baselines failed to reach such MSE. 
    \label{fig:time}}
    \vspace{-2em}

\end{figure}

\subsection{Wall-Time Result}\label{sec:wall-time}

Despite the proposed method's advantages in equal sample comparisons, it is still a question whether such advantage can be translated to wall-time benefits since our method induces significant training and inference overheads. 
For example, even though generating the data required for training takes about 12 minutes, our training procedure takes much longer because it requires computing a higher-order gradient of a neural network for every iteration.
Computing higher-order gradients makes each training iteration slow.
Higher-order gradients also make training less stable, preventing the use of a higher learning rate; as a result, our network takes a large number of iterations to converge.
Our method requires evaluating both the integral and the derivative network during inference.
This means each inference step takes longer than WoS.
Nonetheless, we want to demonstrate that our method can produce wall-time benefits over the baselines in applications that need to produce many very accurate queries.
Specifically, we study where our method has wall-time benefits using the Blub shape in the 3D Laplace experiments (Sec~\ref{sec:result-laplace}).
First, we run all methods to produce a PDE solution of $1024 \times 1024$ resolution to achieve an MSE lower than $3\times 10^{-5}$ using a RTX 2080 Ti GPU.
For each method, we record the detailed compute time breakdown of training and inference time in Figure \ref{fig:time}. 
Including training overhead, our method requires the least amount of time to reach the target accuracy, while other neural CV baselines fail to reach such accuracy within 200 minutes.
While our method spent about half the amount of time in training, our method can reach the same accuracy with a significantly shorter inference time.
This suggests that our method requires fewer walks to reach the same accuracy.
As a result, our method can outperform baselines when it requires a lot of inference samples to reach the target accuracy. 
We also provide results on combining our method with the caching technique \cite{sacache} to demonstrate how such an orthogonal technique can be applied to further improve our method's wall-time efficiency (See Ours+Cache in Figure~\ref{fig:time}).
\begin{table}
  \centering
  \caption{MSE achieved when producing a solution of 3D Laplace equation of the Spot shape in $1024\times 1024$ resolution within 1 hour of inference wall-time. \label{tab:time-profile}\label{tab:time}}
  \vspace{-1.em}
  \begin{tabular}{@{}lcccc@{}}
    \toprule
    Methods & NF & POLY & WoS & Ours  \\ 
    \midrule
    MSE & $2.4 \times 10^{-3}$ & $2.29 \times 10^{-4}$ & $5.45 \times 10^{-5}$ & $2.76 \times 10^{-5}$\\ 
    \bottomrule
  \end{tabular}
\vspace{-1.5em}

\end{table}

Finally, we provide an equal inference time comparison in Table~\ref{tab:time}, where we report the MSE that each method achieves given 1 hour of compute time when creating a $1024\times 1024$ solution image.
With Figure \ref{fig:time}, it demonstrates that our method is faster than baselines in circumstances where a lot of inference samples are needed.

\section{Conclusion and Discussion}
In this paper, we propose a novel method to enable using arbitrary neural network architectures for control variates.
Different from existing methods which mostly deploy a learnable model to approximate the integrand, we ask the neural network to approximate the antiderivative of the integrand instead.
The key insight is that one can use automatic differentiation to derive a network with known integral from the network that approximates the antiderivative.
We apply this idea to reduce the variance of Walk-on-sphere~\cite{WoS} Monte Carlo PDE solvers.
Results suggest that our method is able to create unbiased control variates estimators from various neural network architectures and some of these networks can perform better than all baselines.
\paragraph{Limitations and future works.}
Control variates estimator usually requires more computation for each sampling step because we also need to evaluate in additional $G$ and $g$ for every step.
This suggests that the equal-sample performance improvement might not translate to performance improvement in terms of FLOPs, wall time, or energy.
In more challenging settings where the integrand $f$ or sampling probability $P$ is difficult to evaluate, our proposed approach might provide advantages in wall time.
Computing the integration requires evaluating the antiderivative network $2^d$ times, where $d$ is the integral dimension.
This prevents our method from applying to higher dimensional space.
One potential future direction  to leverage importance sampling to improve training and inference sampling efficiency as demonstrated in \citet{muller2020ncv}.
An other interesting direction is using these neural techniques as carriers to solve inverse problems similar to \citet{nicolet2023recursive}.
\begin{acks}
This research was supported in part by the National Science Foundation under grant 2144117, and by the ARL grant W911NF-21-2-0104.
We want to thank Steve Marschner, Yitong Deng, Wenqi Xian, Rohan Sawhney, and George Nakayama for their feedback.
\end{acks}

\bibliographystyle{ACM-Reference-Format}
\bibliography{main}


\begin{thebibliography}{41}


\ifx \showCODEN    \undefined \def \showCODEN     #1{\unskip}     \fi
\ifx \showDOI      \undefined \def \showDOI       #1{#1}\fi
\ifx \showISBNx    \undefined \def \showISBNx     #1{\unskip}     \fi
\ifx \showISBNxiii \undefined \def \showISBNxiii  #1{\unskip}     \fi
\ifx \showISSN     \undefined \def \showISSN      #1{\unskip}     \fi
\ifx \showLCCN     \undefined \def \showLCCN      #1{\unskip}     \fi
\ifx \shownote     \undefined \def \shownote      #1{#1}          \fi
\ifx \showarticletitle \undefined \def \showarticletitle #1{#1}   \fi
\ifx \showURL      \undefined \def \showURL       {\relax}        \fi
\providecommand\bibfield[2]{#2}
\providecommand\bibinfo[2]{#2}
\providecommand\natexlab[1]{#1}
\providecommand\showeprint[2][]{arXiv:#2}

\bibitem[Bakbouk and Peers(2023)]%
        {bakbouk2023mean}
\bibfield{author}{\bibinfo{person}{Ghada Bakbouk} {and} \bibinfo{person}{Pieter Peers}.} \bibinfo{year}{2023}\natexlab{}.
\newblock \showarticletitle{Mean Value Caching for Walk on Spheres}.
\newblock \bibinfo{journal}{\emph{Eurographics}} (\bibinfo{year}{2023}).
\newblock


\bibitem[Chen et~al\mbox{.}(2018)]%
        {chen2018neural}
\bibfield{author}{\bibinfo{person}{Ricky~TQ Chen}, \bibinfo{person}{Yulia Rubanova}, \bibinfo{person}{Jesse Bettencourt}, {and} \bibinfo{person}{David~K Duvenaud}.} \bibinfo{year}{2018}\natexlab{}.
\newblock \showarticletitle{Neural ordinary differential equations}.
\newblock \bibinfo{journal}{\emph{Advances in neural information processing systems}}  \bibinfo{volume}{31} (\bibinfo{year}{2018}), \bibinfo{pages}{1--9}.
\newblock


\bibitem[Clarberg and Akenine-M{\"o}ller(2008)]%
        {clarberg2008exploiting}
\bibfield{author}{\bibinfo{person}{Petrik Clarberg} {and} \bibinfo{person}{Tomas Akenine-M{\"o}ller}.} \bibinfo{year}{2008}\natexlab{}.
\newblock \showarticletitle{Exploiting visibility correlation in direct illumination}. In \bibinfo{booktitle}{\emph{Computer Graphics Forum}}, Vol.~\bibinfo{volume}{27}. Wiley Online Library, \bibinfo{pages}{1125--1136}.
\newblock


\bibitem[Crane et~al\mbox{.}(2013)]%
        {spot}
\bibfield{author}{\bibinfo{person}{Keenan Crane}, \bibinfo{person}{Ulrich Pinkall}, {and} \bibinfo{person}{Peter Schr{\"o}der}.} \bibinfo{year}{2013}\natexlab{}.
\newblock \showarticletitle{Robust fairing via conformal curvature flow}.
\newblock \bibinfo{journal}{\emph{ACM Transactions on Graphics (TOG)}} \bibinfo{volume}{32}, \bibinfo{number}{4} (\bibinfo{year}{2013}), \bibinfo{pages}{1--10}.
\newblock


\bibitem[Dinh et~al\mbox{.}(2014)]%
        {dinh2014nice}
\bibfield{author}{\bibinfo{person}{Laurent Dinh}, \bibinfo{person}{David Krueger}, {and} \bibinfo{person}{Yoshua Bengio}.} \bibinfo{year}{2014}\natexlab{}.
\newblock \showarticletitle{Nice: Non-linear independent components estimation}, In \bibinfo{booktitle}{International Conference on Learning Representations}.
\newblock \bibinfo{journal}{\emph{International Conference on Learning Representations}}.
\newblock


\bibitem[Dinh et~al\mbox{.}(2016)]%
        {Dinh2016DensityEU}
\bibfield{author}{\bibinfo{person}{Laurent Dinh}, \bibinfo{person}{Jascha Sohl-Dickstein}, {and} \bibinfo{person}{Samy Bengio}.} \bibinfo{year}{2016}\natexlab{}.
\newblock \showarticletitle{Density estimation using Real NVP}. In \bibinfo{booktitle}{\emph{International Conference on Learning Representations}}.
\newblock


\bibitem[Ech-Chafiq et~al\mbox{.}(2021)]%
        {ElFilaliEchChafiq2021AutomaticCV}
\bibfield{author}{\bibinfo{person}{Zineb El~Filali Ech-Chafiq}, \bibinfo{person}{J{\'e}r{\^o}me Lelong}, {and} \bibinfo{person}{Adil Reghai}.} \bibinfo{year}{2021}\natexlab{}.
\newblock \showarticletitle{Automatic control variates for option pricing using neural networks}.
\newblock \bibinfo{journal}{\emph{Monte Carlo Methods and Applications}}  \bibinfo{volume}{27} (\bibinfo{year}{2021}), \bibinfo{pages}{91 -- 104}.
\newblock
\urldef\tempurl%
\url{https://api.semanticscholar.org/CorpusID:234204906}
\showURL{%
\tempurl}


\bibitem[Geffner and Domke(2018)]%
        {Geffner2018UsingLE}
\bibfield{author}{\bibinfo{person}{Tomas Geffner} {and} \bibinfo{person}{Justin Domke}.} \bibinfo{year}{2018}\natexlab{}.
\newblock \showarticletitle{Using large ensembles of control variates for variational inference}.
\newblock \bibinfo{journal}{\emph{Advances in Neural Information Processing Systems}}  \bibinfo{volume}{31} (\bibinfo{year}{2018}).
\newblock


\bibitem[Glynn and Szechtman(2002)]%
        {glynn2002some}
\bibfield{author}{\bibinfo{person}{Peter~W Glynn} {and} \bibinfo{person}{Roberto Szechtman}.} \bibinfo{year}{2002}\natexlab{}.
\newblock \showarticletitle{Some new perspectives on the method of control variates}. In \bibinfo{booktitle}{\emph{Monte Carlo and Quasi-Monte Carlo Methods 2000: Proceedings of a Conference held at Hong Kong Baptist University, Hong Kong SAR, China, November 27--December 1, 2000}}. Springer, \bibinfo{pages}{27--49}.
\newblock


\bibitem[Hua et~al\mbox{.}(2023)]%
        {hua2023revisiting}
\bibfield{author}{\bibinfo{person}{Qingqin Hua}, \bibinfo{person}{Pascal Grittmann}, {and} \bibinfo{person}{Philipp Slusallek}.} \bibinfo{year}{2023}\natexlab{}.
\newblock \showarticletitle{Revisiting controlled mixture sampling for rendering applications}.
\newblock \bibinfo{journal}{\emph{ACM Transactions on Graphics (TOG)}} \bibinfo{volume}{42}, \bibinfo{number}{4} (\bibinfo{year}{2023}), \bibinfo{pages}{1--13}.
\newblock


\bibitem[Kingma and Ba(2014)]%
        {kingma2014adam}
\bibfield{author}{\bibinfo{person}{Diederik~P. Kingma} {and} \bibinfo{person}{Jimmy Ba}.} \bibinfo{year}{2014}\natexlab{}.
\newblock \showarticletitle{Adam: A Method for Stochastic Optimization}.
\newblock \bibinfo{journal}{\emph{International Conference on Learning Representations}}  \bibinfo{volume}{abs/1412.6980} (\bibinfo{year}{2014}).
\newblock
\urldef\tempurl%
\url{https://api.semanticscholar.org/CorpusID:6628106}
\showURL{%
\tempurl}


\bibitem[Kn\"{o}ppel et~al\mbox{.}(2015)]%
        {blub}
\bibfield{author}{\bibinfo{person}{Felix Kn\"{o}ppel}, \bibinfo{person}{Keenan Crane}, \bibinfo{person}{Ulrich Pinkall}, {and} \bibinfo{person}{Peter Schr\"{o}der}.} \bibinfo{year}{2015}\natexlab{}.
\newblock \showarticletitle{Stripe Patterns on Surfaces}.
\newblock \bibinfo{journal}{\emph{ACM Trans. Graph.}}  \bibinfo{volume}{34} (\bibinfo{year}{2015}).
\newblock
Issue 4.


\bibitem[Kutz et~al\mbox{.}(2017)]%
        {kutz2017spectral}
\bibfield{author}{\bibinfo{person}{Peter Kutz}, \bibinfo{person}{Ralf Habel}, \bibinfo{person}{Yining~Karl Li}, {and} \bibinfo{person}{Jan Nov{\'a}k}.} \bibinfo{year}{2017}\natexlab{}.
\newblock \showarticletitle{Spectral and decomposition tracking for rendering heterogeneous volumes}.
\newblock \bibinfo{journal}{\emph{ACM Transactions on Graphics (TOG)}} \bibinfo{volume}{36}, \bibinfo{number}{4} (\bibinfo{year}{2017}), \bibinfo{pages}{1--16}.
\newblock


\bibitem[Lafortune and Willems(1994)]%
        {lafortune1994using}
\bibfield{author}{\bibinfo{person}{Eric~P Lafortune} {and} \bibinfo{person}{Yves~D Willems}.} \bibinfo{year}{1994}\natexlab{}.
\newblock \bibinfo{booktitle}{\emph{Using the modified phong reflectance model for physically based rendering}}.
\newblock \bibinfo{publisher}{Katholieke Universiteit Leuven. Departement Computerwetenschappen}.
\newblock


\bibitem[Li et~al\mbox{.}(2023)]%
        {sacache}
\bibfield{author}{\bibinfo{person}{Zilu Li}, \bibinfo{person}{Guandao Yang}, \bibinfo{person}{Xi Deng}, \bibinfo{person}{Christopher De~Sa}, \bibinfo{person}{Bharath Hariharan}, {and} \bibinfo{person}{Steve Marschner}.} \bibinfo{year}{2023}\natexlab{}.
\newblock \showarticletitle{Neural Caches for Monte Carlo Partial Differential Equation Solvers}. In \bibinfo{booktitle}{\emph{SIGGRAPH Asia 2023 Conference Papers}} \emph{(\bibinfo{series}{SA '23})}. \bibinfo{publisher}{Association for Computing Machinery}, \bibinfo{address}{New York, NY, USA}, Article \bibinfo{articleno}{34}, \bibinfo{numpages}{10}~pages.
\newblock
\showISBNx{9798400703157}
\urldef\tempurl%
\url{https://doi.org/10.1145/3610548.3618141}
\showDOI{\tempurl}


\bibitem[Lindell et~al\mbox{.}(2021)]%
        {lindell2021autoint}
\bibfield{author}{\bibinfo{person}{David~B Lindell}, \bibinfo{person}{Julien~NP Martel}, {and} \bibinfo{person}{Gordon Wetzstein}.} \bibinfo{year}{2021}\natexlab{}.
\newblock \showarticletitle{Autoint: Automatic integration for fast neural volume rendering}. In \bibinfo{booktitle}{\emph{Proceedings of the IEEE/CVF Conference on Computer Vision and Pattern Recognition}}. \bibinfo{pages}{14556--14565}.
\newblock


\bibitem[Loh(1995)]%
        {loh1995method}
\bibfield{author}{\bibinfo{person}{Wing~Wah Loh}.} \bibinfo{year}{1995}\natexlab{}.
\newblock \bibinfo{booktitle}{\emph{On the method of control variates}}.
\newblock \bibinfo{publisher}{Stanford University}.
\newblock


\bibitem[Ma{\^\i}tre and Santos-Mateos(2023)]%
        {maitre2023multi}
\bibfield{author}{\bibinfo{person}{Daniel Ma{\^\i}tre} {and} \bibinfo{person}{Roi Santos-Mateos}.} \bibinfo{year}{2023}\natexlab{}.
\newblock \showarticletitle{Multi-variable integration with a neural network}.
\newblock \bibinfo{journal}{\emph{Journal of High Energy Physics}} \bibinfo{volume}{2023}, \bibinfo{number}{3} (\bibinfo{year}{2023}), \bibinfo{pages}{1--16}.
\newblock


\bibitem[Mehta et~al\mbox{.}(2021)]%
        {mehta2021modulated}
\bibfield{author}{\bibinfo{person}{Ishit Mehta}, \bibinfo{person}{Micha{\"e}l Gharbi}, \bibinfo{person}{Connelly Barnes}, \bibinfo{person}{Eli Shechtman}, \bibinfo{person}{Ravi Ramamoorthi}, {and} \bibinfo{person}{Manmohan Chandraker}.} \bibinfo{year}{2021}\natexlab{}.
\newblock \showarticletitle{Modulated periodic activations for generalizable local functional representations}. In \bibinfo{booktitle}{\emph{Proceedings of the IEEE/CVF International Conference on Computer Vision}}. \bibinfo{pages}{14214--14223}.
\newblock


\bibitem[Miller et~al\mbox{.}(2023)]%
        {B-cache}
\bibfield{author}{\bibinfo{person}{Bailey Miller}, \bibinfo{person}{Rohan Sawhney}, \bibinfo{person}{Keenan Crane}, {and} \bibinfo{person}{Ioannis Gkioulekas}.} \bibinfo{year}{2023}\natexlab{}.
\newblock \showarticletitle{Boundary Value Caching for Walk on Spheres}.
\newblock \bibinfo{journal}{\emph{ACM Trans. Graph.}} \bibinfo{volume}{42}, \bibinfo{number}{4}, Article \bibinfo{articleno}{82} (\bibinfo{date}{jul} \bibinfo{year}{2023}), \bibinfo{numpages}{11}~pages.
\newblock
\showISSN{0730-0301}
\urldef\tempurl%
\url{https://doi.org/10.1145/3592400}
\showDOI{\tempurl}


\bibitem[M\"uller et~al\mbox{.}(2022)]%
        {mueller2022instant}
\bibfield{author}{\bibinfo{person}{Thomas M\"uller}, \bibinfo{person}{Alex Evans}, \bibinfo{person}{Christoph Schied}, {and} \bibinfo{person}{Alexander Keller}.} \bibinfo{year}{2022}\natexlab{}.
\newblock \showarticletitle{Instant Neural Graphics Primitives with a Multiresolution Hash Encoding}.
\newblock \bibinfo{journal}{\emph{ACM Trans. Graph.}} \bibinfo{volume}{41}, \bibinfo{number}{4}, Article \bibinfo{articleno}{102} (\bibinfo{date}{July} \bibinfo{year}{2022}), \bibinfo{numpages}{15}~pages.
\newblock
\urldef\tempurl%
\url{https://doi.org/10.1145/3528223.3530127}
\showDOI{\tempurl}


\bibitem[M{\"u}ller et~al\mbox{.}(2020)]%
        {muller2020ncv}
\bibfield{author}{\bibinfo{person}{Thomas M{\"u}ller}, \bibinfo{person}{Fabrice Rousselle}, \bibinfo{person}{Alexander Keller}, {and} \bibinfo{person}{Jan Nov{\'a}k}.} \bibinfo{year}{2020}\natexlab{}.
\newblock \showarticletitle{Neural control variates}.
\newblock \bibinfo{journal}{\emph{ACM Transactions on Graphics (TOG)}} \bibinfo{volume}{39}, \bibinfo{number}{6} (\bibinfo{year}{2020}), \bibinfo{pages}{1--19}.
\newblock


\bibitem[Nicolet et~al\mbox{.}(2023)]%
        {nicolet2023recursive}
\bibfield{author}{\bibinfo{person}{Baptiste Nicolet}, \bibinfo{person}{Fabrice Rousselle}, \bibinfo{person}{Jan Novak}, \bibinfo{person}{Alexander Keller}, \bibinfo{person}{Wenzel Jakob}, {and} \bibinfo{person}{Thomas M{\"u}ller}.} \bibinfo{year}{2023}\natexlab{}.
\newblock \showarticletitle{Recursive control variates for inverse rendering}.
\newblock \bibinfo{journal}{\emph{ACM Transactions on Graphics (TOG)}} \bibinfo{volume}{42}, \bibinfo{number}{4} (\bibinfo{year}{2023}), \bibinfo{pages}{1--13}.
\newblock


\bibitem[Norman and Moore(1977)]%
        {norman1977implementing}
\bibfield{author}{\bibinfo{person}{Arthur~C Norman} {and} \bibinfo{person}{PMA Moore}.} \bibinfo{year}{1977}\natexlab{}.
\newblock \showarticletitle{Implementing the new Risch integration algorithm}. In \bibinfo{booktitle}{\emph{Proceedings of the 4th International Colloquium on Advanced Computing Methods in Theoretical Physics}}. \bibinfo{pages}{99--110}.
\newblock


\bibitem[Nsampi et~al\mbox{.}(2023)]%
        {nsampi2023neural}
\bibfield{author}{\bibinfo{person}{Ntumba~Elie Nsampi}, \bibinfo{person}{Adarsh Djeacoumar}, \bibinfo{person}{Hans-Peter Seidel}, \bibinfo{person}{Tobias Ritschel}, {and} \bibinfo{person}{Thomas Leimk{\"u}hler}.} \bibinfo{year}{2023}\natexlab{}.
\newblock \showarticletitle{Neural Field Convolutions by Repeated Differentiation}.
\newblock \bibinfo{journal}{\emph{ACM Transactions on Graphics (TOG)}} \bibinfo{volume}{42}, \bibinfo{number}{6} (\bibinfo{year}{2023}), \bibinfo{pages}{1--11}.
\newblock


\bibitem[Pajot et~al\mbox{.}(2014)]%
        {pajot2014globally}
\bibfield{author}{\bibinfo{person}{Anthony Pajot}, \bibinfo{person}{Lo{\"\i}c Barthe}, {and} \bibinfo{person}{Mathias Paulin}.} \bibinfo{year}{2014}\natexlab{}.
\newblock \showarticletitle{Globally Adaptive Control Variate for Robust Numerical Integration}.
\newblock \bibinfo{journal}{\emph{SIAM Journal on Scientific Computing}} \bibinfo{volume}{36}, \bibinfo{number}{4} (\bibinfo{year}{2014}), \bibinfo{pages}{A1708--A1730}.
\newblock


\bibitem[Petrosyan et~al\mbox{.}(2020)]%
        {petrosyan2020neural}
\bibfield{author}{\bibinfo{person}{Armenak Petrosyan}, \bibinfo{person}{Anton Dereventsov}, {and} \bibinfo{person}{Clayton~G Webster}.} \bibinfo{year}{2020}\natexlab{}.
\newblock \showarticletitle{Neural network integral representations with the ReLU activation function}. In \bibinfo{booktitle}{\emph{Mathematical and Scientific Machine Learning}}. PMLR, \bibinfo{pages}{128--143}.
\newblock


\bibitem[Qi et~al\mbox{.}(2022)]%
        {qi2022bidirectional}
\bibfield{author}{\bibinfo{person}{Yang Qi}, \bibinfo{person}{Dario Seyb}, \bibinfo{person}{Benedikt Bitterli}, {and} \bibinfo{person}{Wojciech Jarosz}.} \bibinfo{year}{2022}\natexlab{}.
\newblock \showarticletitle{A bidirectional formulation for Walk on Spheres}. In \bibinfo{booktitle}{\emph{Computer Graphics Forum}}, Vol.~\bibinfo{volume}{41}. Wiley Online Library, \bibinfo{pages}{51--62}.
\newblock


\bibitem[Rioux-Lavoie et~al\mbox{.}(2022)]%
        {MCFluid}
\bibfield{author}{\bibinfo{person}{Damien Rioux-Lavoie}, \bibinfo{person}{Ryusuke Sugimoto}, \bibinfo{person}{T{\"u}may {\"O}zdemir}, \bibinfo{person}{Naoharu~H Shimada}, \bibinfo{person}{Christopher Batty}, \bibinfo{person}{Derek Nowrouzezahrai}, {and} \bibinfo{person}{Toshiya Hachisuka}.} \bibinfo{year}{2022}\natexlab{}.
\newblock \showarticletitle{A Monte Carlo Method for Fluid Simulation}.
\newblock \bibinfo{journal}{\emph{ACM Trans. Graph.}} \bibinfo{volume}{41}, \bibinfo{number}{6} (\bibinfo{date}{Nov.} \bibinfo{year}{2022}), \bibinfo{pages}{1--16}.
\newblock


\bibitem[Risch(1969)]%
        {Risch1969ThePO}
\bibfield{author}{\bibinfo{person}{Robert~H. Risch}.} \bibinfo{year}{1969}\natexlab{}.
\newblock \showarticletitle{The problem of integration in finite terms}.
\newblock \bibinfo{journal}{\emph{Trans. Amer. Math. Soc.}}  \bibinfo{volume}{139} (\bibinfo{year}{1969}), \bibinfo{pages}{167--189}.
\newblock
\urldef\tempurl%
\url{https://api.semanticscholar.org/CorpusID:122648356}
\showURL{%
\tempurl}


\bibitem[Rousselle et~al\mbox{.}(2016)]%
        {rousselle2016image}
\bibfield{author}{\bibinfo{person}{Fabrice Rousselle}, \bibinfo{person}{Wojciech Jarosz}, {and} \bibinfo{person}{Jan Nov{\'a}k}.} \bibinfo{year}{2016}\natexlab{}.
\newblock \showarticletitle{Image-space control variates for rendering}.
\newblock \bibinfo{journal}{\emph{ACM Transactions on Graphics (TOG)}} \bibinfo{volume}{35}, \bibinfo{number}{6} (\bibinfo{year}{2016}), \bibinfo{pages}{1--12}.
\newblock


\bibitem[Sala{\"u}n et~al\mbox{.}(2022)]%
        {salaun2022regression}
\bibfield{author}{\bibinfo{person}{Corentin Sala{\"u}n}, \bibinfo{person}{Adrien Gruson}, \bibinfo{person}{Binh-Son Hua}, \bibinfo{person}{Toshiya Hachisuka}, {and} \bibinfo{person}{Gurprit Singh}.} \bibinfo{year}{2022}\natexlab{}.
\newblock \showarticletitle{Regression-based Monte Carlo integration}.
\newblock \bibinfo{journal}{\emph{ACM Transactions on Graphics (TOG)}} \bibinfo{volume}{41}, \bibinfo{number}{4} (\bibinfo{year}{2022}), \bibinfo{pages}{1--14}.
\newblock


\bibitem[Sawhney and Crane(2020)]%
        {WoS}
\bibfield{author}{\bibinfo{person}{Rohan Sawhney} {and} \bibinfo{person}{Keenan Crane}.} \bibinfo{year}{2020}\natexlab{}.
\newblock \showarticletitle{Monte Carlo geometry processing}.
\newblock \bibinfo{journal}{\emph{ACM Trans. Graph.}} \bibinfo{volume}{39}, \bibinfo{number}{4} (\bibinfo{date}{Aug.} \bibinfo{year}{2020}).
\newblock


\bibitem[Sawhney et~al\mbox{.}(2023)]%
        {WoSt}
\bibfield{author}{\bibinfo{person}{Rohan Sawhney}, \bibinfo{person}{Bailey Miller}, \bibinfo{person}{Ioannis Gkioulekas}, {and} \bibinfo{person}{Keenan Crane}.} \bibinfo{year}{2023}\natexlab{}.
\newblock \showarticletitle{Walk on Stars: A Grid-Free Monte Carlo Method for PDEs with Neumann Boundary Conditions}.
\newblock \bibinfo{journal}{\emph{ACM Trans. Graph.}} \bibinfo{volume}{42}, \bibinfo{number}{4}, Article \bibinfo{articleno}{80} (\bibinfo{date}{jul} \bibinfo{year}{2023}), \bibinfo{numpages}{20}~pages.
\newblock
\showISSN{0730-0301}
\urldef\tempurl%
\url{https://doi.org/10.1145/3592398}
\showDOI{\tempurl}


\bibitem[Sawhney et~al\mbox{.}(2022)]%
        {VWos}
\bibfield{author}{\bibinfo{person}{Rohan Sawhney}, \bibinfo{person}{Dario Seyb}, \bibinfo{person}{Wojciech Jarosz}, {and} \bibinfo{person}{Keenan Crane}.} \bibinfo{year}{2022}\natexlab{}.
\newblock \showarticletitle{Grid-free Monte Carlo for PDEs with spatially varying coefficients}.
\newblock \bibinfo{journal}{\emph{ACM Trans. Graph.}} \bibinfo{volume}{41}, \bibinfo{number}{4}, Article \bibinfo{articleno}{53} (\bibinfo{date}{jul} \bibinfo{year}{2022}), \bibinfo{numpages}{17}~pages.
\newblock
\showISSN{0730-0301}
\urldef\tempurl%
\url{https://doi.org/10.1145/3528223.3530134}
\showDOI{\tempurl}


\bibitem[Sitzmann et~al\mbox{.}(2020)]%
        {sitzmann2020implicit}
\bibfield{author}{\bibinfo{person}{Vincent Sitzmann}, \bibinfo{person}{Julien Martel}, \bibinfo{person}{Alexander Bergman}, \bibinfo{person}{David Lindell}, {and} \bibinfo{person}{Gordon Wetzstein}.} \bibinfo{year}{2020}\natexlab{}.
\newblock \showarticletitle{Implicit neural representations with periodic activation functions}.
\newblock \bibinfo{journal}{\emph{Advances in neural information processing systems}}  \bibinfo{volume}{33} (\bibinfo{year}{2020}), \bibinfo{pages}{7462--7473}.
\newblock


\bibitem[Subr(2021)]%
        {subr2021q}
\bibfield{author}{\bibinfo{person}{Kartic Subr}.} \bibinfo{year}{2021}\natexlab{}.
\newblock \showarticletitle{Q-NET: A Network for Low-dimensional Integrals of Neural Proxies}. In \bibinfo{booktitle}{\emph{Computer Graphics Forum}}, Vol.~\bibinfo{volume}{40}. Wiley Online Library, \bibinfo{pages}{61--71}.
\newblock


\bibitem[Tabak and Turner(2013)]%
        {tabak2013family}
\bibfield{author}{\bibinfo{person}{Esteban~G Tabak} {and} \bibinfo{person}{Cristina~V Turner}.} \bibinfo{year}{2013}\natexlab{}.
\newblock \showarticletitle{A family of nonparametric density estimation algorithms}.
\newblock \bibinfo{journal}{\emph{Communications on Pure and Applied Mathematics}} \bibinfo{volume}{66}, \bibinfo{number}{2} (\bibinfo{year}{2013}), \bibinfo{pages}{145--164}.
\newblock


\bibitem[Veach(1998)]%
        {veach1998robust}
\bibfield{author}{\bibinfo{person}{Eric Veach}.} \bibinfo{year}{1998}\natexlab{}.
\newblock \bibinfo{booktitle}{\emph{Robust Monte Carlo methods for light transport simulation}}.
\newblock \bibinfo{publisher}{Stanford University}.
\newblock


\bibitem[Wan et~al\mbox{.}(2020)]%
        {Wan2019NeuralCV}
\bibfield{author}{\bibinfo{person}{Ruosi Wan}, \bibinfo{person}{Mingjun Zhong}, \bibinfo{person}{Haoyi Xiong}, {and} \bibinfo{person}{Zhanxing Zhu}.} \bibinfo{year}{2020}\natexlab{}.
\newblock \showarticletitle{Neural control variates for Monte Carlo variance reduction}. In \bibinfo{booktitle}{\emph{Machine Learning and Knowledge Discovery in Databases: European Conference, ECML PKDD 2019, W{\"u}rzburg, Germany, September 16--20, 2019, Proceedings, Part II}}. Springer, \bibinfo{pages}{533--547}.
\newblock


\bibitem[Zhou and Yu(2023)]%
        {zhou2023automatic}
\bibfield{author}{\bibinfo{person}{Zihao Zhou} {and} \bibinfo{person}{Rose Yu}.} \bibinfo{year}{2023}\natexlab{}.
\newblock \showarticletitle{Automatic Integration for Fast and Interpretable Neural Point Processes}. In \bibinfo{booktitle}{\emph{Learning for Dynamics and Control Conference}}. PMLR, \bibinfo{address}{University of Pennsylvania}, \bibinfo{pages}{573--585}.
\newblock


\end{thebibliography}

\end{document}